# Efficient Fine-Tuning of Large Language Models for Automated Medical Documentation

1st *Hui Yi, Leong
*Data Science Institute,*
*University of Chicago, Chicago, Illinois, USA.*

2nd Yi Fan, Gao
*Data Science Institute,*
*University of Chicago, Chicago, Illinois, USA.*

3rd Shuai, Ji
*Data Science Institute,*
*University of Chicago, Chicago, Illinois, USA.*

4th Yang Zhang
*Metropolitan College,*
*Boston University, Boston MA, USA.*

5th Uktu Pamuksuz
*Data Science Institute,*
*University of Chicago, Chicago, Illinois, USA.*

*Abstract*— **Scientific research indicates that for every hour spent in direct patient care, physicians spend nearly two additional hours on administrative tasks, particularly on electronic health records (EHRs) and desk work. This excessive administrative burden not only reduces the time available for patient care but also contributes to physician burnout and inefficiencies in healthcare delivery. To address these challenges, this study introduces MediGen, a fine-tuned large language model (LLM) designed to automate the generation of medical reports from medical dialogues. By leveraging state-of-the-art methodologies for fine-tuning open-source pretrained models, including LLaMA3-8B, MediGen achieves high accuracy in transcribing and summarizing clinical interactions. The fine-tuned LLaMA3-8B model demonstrated promising results, achieving a ROUGE score of 58% and a BERTScore-F1 of 72%, indicating its effectiveness in generating accurate and clinically relevant medical reports. These findings suggest that MediGen has the potential to significantly reduce the administrative workload on physicians, improving both healthcare efficiency and physician well-being.**

## I. INTRODUCTION

The integration of Artificial Intelligence (AI) into healthcare has led to significant advancements in areas such as disease diagnosis, treatment planning, and medical reporting [1]. In particular, Large Language Models (LLMs) have demonstrated the potential to transform medical documentation by automating the generation of clinical notes. However, despite the progress made in electronic health records (EHRs) systems, clinicians still face the challenge of excessive administrative burdens, which can reduce the time available for patient care and contribute to physician burnout. Studies suggest that for every hour of clinical work, physicians spend nearly two additional hours managing EHRs and other administrative tasks. This clerical burden contributes to an increased risk of professional burnout [2]. This imbalance hampers the efficiency of healthcare delivery and impacts both patient outcomes and physician well-being.

To address these challenges, this study proposes a novel approach to automating the generation of clinical reports from medical dialogues. By leveraging a fine-tuned Large Language Model, the goal is to reduce the time physicians spend on documentation, allowing them to focus more on patient care. The model is designed to accurately transcribe and summarize medical interactions into structured SOAP (Subjective, Objective, Assessment, and Plan) notes. Our fine-tuned LLaMA3-8B model, MediGen, has shown promising results in improving content accuracy and efficiency, positioning it as a viable tool for enhancing medical workflows.

This paper details the methodology for fine-tuning LLMs to generate medical reports and discusses the implications of this technology for improving healthcare efficiency and reducing physician burnout. Our approach emphasizes training and optimizing models to generate medical notes with high accuracy while minimizing computational resource requirements, laying the groundwork for further advancements in automated medical documentation.

## II. RELATED WORK

The increasing administrative burden associated with medical documentation has been a growing concern in healthcare. Several studies have explored the use of Artificial Intelligence (AI) and Natural Language Processing (NLP) to automate clinical note generation, aiming to alleviate this workload. The emergence of Large Language Models (LLMs) has significantly advanced this field, providing new methods for summarizing medical dialogues and automating the documentation process.

One of the earliest approaches to automating clinical note generation involved the use of Recurrent Neural Networks (RNNs), which were effective in handling sequential data such as conversations. These models predict subsequent words based on the preceding context, making them suitable for speech recognition tasks. However, RNNs had significant limitations, particularly their inability to handle long-range dependencies effectively. This shortcoming often led to incomplete or inaccurate summaries in longer dialogues, reducing their utility in real-world medical settings.

The development of transformer-based models, particularly those utilizing attention mechanisms, revolutionized NLP by allowing models to weigh the importance of words regardless of their position in a sequence. Transformers, such as BERT and GPT, dramatically improved the accuracy of medical report generation. However, high computational cost and memory requirements present significant barriers to widespread

* *Corresponding author: yuki.leong@uchicago.edu*

implementation, especially in resource-limited healthcare environments [3].

Several works have focused on leveraging LLMs for medical report generation. For instance, Yim et al. developed the ACI-BENCH dataset to benchmark systems for clinical note generation from doctor-patient dialogues [4]. While this dataset has provided a valuable resource for evaluating model performance, it is limited in scope due to its relatively small size and lack of diversity in clinical scenarios. This limitation restricts models' ability to generalize across various medical contexts and patient populations, potentially leading to inaccuracies in less represented cases.

## III. METHODOLOGY

The Figure 1 describes the training pipeline for the MediGen model, including dataset preparation, data preprocessing, model selection, and fine-tuning techniques. The evaluation results will be discussed later.

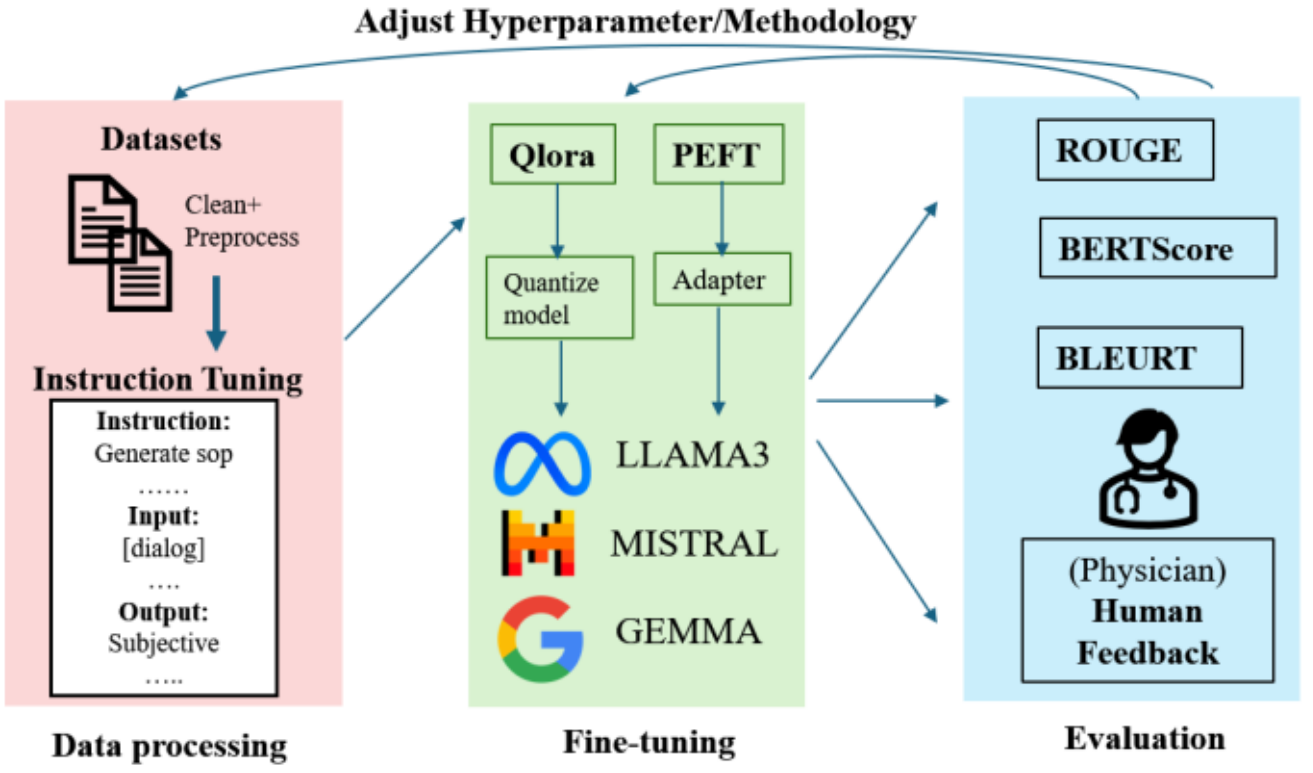

Fig. 1. Overview of the MediGen Model Training and Evaluation Pipeline

### A. Dataset

For training and fine-tuning our model, we utilized the ACI-BENCH dataset [4], a publicly available medical dialogue corpus. Acquiring real-time medical datasets poses significant challenges, as medical conversations are rarely recorded, and privacy laws and ethical considerations make data sharing difficult. Despite its size limitations, the ACI-BENCH dataset is currently the largest publicly accessible resource for model-assisted clinical note generation. It encompasses various clinical scenarios, including virtual assistants, virtual scribes, and natural patient-doctor interactions, making it an essential benchmark for this task.

The ACI-BENCH dataset contains 16MB of data, comprising 207 doctor-patient role-played dialogue-note pairs, as shown in Table 1. Each dialogue averages 1,302 tokens, with the corresponding SOAP notes averaging 490 tokens. The dataset includes a wide range of clinical scenarios, from virtual assistant calls to real-life doctor-patient exchanges, offering a diverse training base that enhances the model's ability to generalize. The dataset was divided into three subsets: 67 dialogues for training, 20 for validation, and 120 for testing (further divided into three test sets). This structured split enables a comprehensive evaluation of the model's performance on unseen data, ensuring a robust assessment of its generalization capabilities..

Table 1. Statistics of the ACI-BENCH dataset.

| | Train | Valid | Test1 | Test2 | Test3 |
|---|---|---|---|---|---|
| Number encounters | 67 | 20 | 40 | 40 | 40 |
| *Dialogue* | | | | | |
| Avg number turns | 56 | 53 | 52 | 56 | 58 |
| Avg length (tok) | 1301 | 1221 | 1231 | 1382 | 1334 |

### B. Data Preprocessing

Preprocessing was a crucial step to ensure the quality and consistency of the input data. The dialogues were cleaned by removing irrelevant or repeated words, correcting punctuation, and tokenizing the text using an instruction-tuning format. Each dialogue was paired with its corresponding SOAP note and segmented into input-output pairs for training.

We employed the following steps during preprocessing:

**Text normalization**: All text data was standardized by removing special characters and ensuring consistency in medical terminology.

**Tokenization**: The dialogues and notes were tokenized using a sentencepiece tokenizer pre-trained on medical texts.

**Speaker identification**: We identified speaker roles (doctor or patient) to ensure that the model could correctly distinguish between medical advice and patient symptoms or queries.

### C. Model Selection

Given the limitations of **RNNs** in handling long-range dependencies, we opted for transformer-based models that utilize attention mechanisms to retain critical information across lengthy medical dialogues. While transformer models like **BERT** and **GPT** are effective, they come with significant computational and memory requirements, making them less practical for resource-constrained healthcare settings [3]. This is especially important in medical dialogue summarization, where the model must efficiently process long conversations and scale effectively.

We selected **LLaMA3-8B** [5] as it balances performance and resource efficiency. It is more resource-friendly and easier to fine-tune using advanced techniques like **QLoRA** [6] and **PEFT** [7] , which reduce memory and computational overhead without sacrificing performance. These features make **LLaMA3-8B** [5] ideal for healthcare, especially in handling long medical conversations. Additionally, we included **GEMMA-7B** [8] and **Mistral-7B** [9] for ablation studies to assess how different architectures impact summarization performance, providing valuable insights for optimizing models for healthcare tasks.

### D. Fine-Tuning Techniques

To optimize MediGen's performance while addressing computational constraints, we employed three advanced fine-tuning strategies: Quantized Low-Rank Adaptation (QLoRA) [5], Parameter-Efficient Fine-Tuning (PEFT) [6], and Instruction Tuning. These techniques were selected to enhance the model's ability to generate accurate and detailed SOAP notes from medical dialogues while reducing memory usage and computational load.

*a)* ***Quantized Low-Rank Adaptation (QLoRA):*** We selected QLoRA to reduce memory consumption by quantizing model parameters to 4 bits and applying low-rank adaptations. This allows efficient fine-tuning with fewer resources, crucial for real-world healthcare settings where computational power may be limited, without sacrificing the performance of the original model.

*b)* ***Parameter Efficient Fine-Tuning (PEFT):*** PEFT was chosen because it allows the fine-tuning of only a small subset of the model's parameters, particularly those most critical for the task. This significantly reduces the number of trainable parameters, making the process more computationally efficient while maintaining accuracy—an essential consideration for scalable healthcare applications.

*c)* ***Instruction Tuning Method:*** Transformer models often struggle with maintaining nuanced details in medical conversations, which can lead to overly generic summaries that miss vital information. To address this, we employed instruction tuning, a technique that explicitly directs the model to organize information into specific SOAP note sections (e.g., Subjective, Objective, Assessment, Plan). This method ensures that the model can generate coherent, accurate, and actionable medical summaries by following structured input-output pairs. Figure 2 illustrates the instruction tuning format for generating SOAP notes.

| **Instruction :** | **Input:** | **Output:** |
|---|---|---|
| Summarize medical dialogues into a SOAP note format, where the note is divided into four continuous sections: SUBJECTIVE, OBJECTIVE_EXAM ,OBJECTIVE_RESULTS, and ASSESSMENT_AND_PLAN. The SUBJECTIVE section should contain information from the verbal examination..... | [doctor] hi , bryan . how are you ?<br>[patient] i'm doing well . i'm a little sore .<br>[doctor] so bryan is a 55-year-old male with a past medical history significant for prior discectomy , who presents with back pain . so , bryan , what happened to your back ?<br>[patient] you ... my wife made me push a , uh , refrigerator out through the other room , and when i was helping to move it , i felt something in my back on the lower right side .<br>[doctor] okay , on the lower right side of this back ?<br>[patient] yes .<br>[doctor] okay . those wives , always making you do stuff .<br>[patient] yes .<br>[doctor] and what day did this happen on ? how long ago ?<br>[patient] uh , this was about five days ago .<br>.... | **CHIEF COMPLAINT**<br>Back pain.<br><br>**HISTORY OF PRESENT ILLNESS**<br>Bryan Smith is a 55-year-old male with a past medical history significant for and prior discectomy, who presents with back pain.<br><br>**REVIEW OF SYSTEMS**<br>.....<br><br>**PHYSICAL EXAMINATION**<br>.....<br><br>**RESULTS**<br>.....<br><br>**ASSESSMENT AND PLAN**<br>..... |

Figure 2: Instruction Tuning Format for Generating SOAP Notes

*d)* ***Adjust hyperparameters****:* Hyperparameters were adjusted based on iterative evaluations with healthcare providers. By involving real users in the evaluation process, we continuously refined the model's performance to ensure accuracy and prevent overfitting.

## IV. EXPERRIMENT RESULTS

### *A. Evaluation Results*

We used a combination of quantitative metrics (ROUGE, BERTScore, and BLEURT) and qualitative assessments to evaluate the performance of MediGen, as shown in Table 2. The quantitative metrics focused on the accuracy, relevance, and coherence of the generated medical reports, while the qualitative assessments involved expert reviews by medical professionals. The fine-tuned MediGen model, based on LLaMA3-8B, was evaluated using the ACI-BENCH dataset through three test sets, and the results were averaged to ensure consistency across diverse doctor-patient dialogues.

Table 2. Performance Evaluation of MediGen LLM (LLaMA3-8B-FT) Against Baseline Models

| | Rouge1 | Rouge2 | Rouge Lsum | BERTScore-F1 | BLEURT |
|---|---|---|---|---|---|
| BART+FTSAMsum | 52.77 | **24.61** | 47.94 | 68.42 | 37.41 |
| GPT4o | 50.64 | 21.1 | 47.64 | 65.89 | 39.65 |
| Phi-3-mini-4k-instruct | 18.54 | 1.37 | 16.83 | 48.87 | 31.38 |
| Gemma-7b | 10.11 | 3.11 | 9.14 | 34.44 | 31.23 |
| Mistral-7b | 20.32 | 9.9 | 18.82 | 45.52 | 28.91 |
| Mistral-7b-instruct | 30.44 | 12.95 | 27.7 | 51.05 | 35.88 |
| Llama3-8b | 30.29 | 11.42 | 27.42 | 54.41 | 35.59 |
| Llama3-8b-instruct | 29.61 | 11.17 | 27.43 | 54.25 | 31.35 |
| Llama3-8b-FT(Ours) | **68.22** | 24.59 | **53.84** | **72.1** | **41.15** |

**ROUGE (Recall-Oriented Understudy for Gisting Evaluation)** measures the overlap of n-grams between generated notes and reference notes. MediGen achieved an average ROUGE-1 score of 58.22% and ROUGE-Lsum of 53.84%, surpassing the current leading model on ACI-Bench, BART+FTSMSum, which scored 52.64% for ROUGE-1 and 47.94% for ROUGE-Lsum. MediGen also performed similarly to BART+FTSMSum on ROUGE-2, demonstrating strong accuracy in capturing key medical details from dialogues.

The **BERTScore** metric, which assesses the semantic similarity between generated and reference texts, was used to further evaluate the quality of the notes. MediGen achieved a BERTScore-F1 of 72.1%, reflecting strong semantic relevance and coherence in its summaries.

When compared to traditional metrics like ROUGE and BERTScore, the **BLEURT** results provided a more nuanced view of the model's ability to generate clinically relevant content. For example, while most models performed similarly on BLEUR, BLEURT revealed that MediGen consistently captured subtle differences in meaning better than all other models, particularly in more complex medical dialogues, achieves a BLEU score of 41,15%.

For qualitative evaluation, we conducted a clinical review where medical professionals assessed the generated notes for accuracy, completeness, and clinical relevance. In collaboration with the University of Chicago Medical Center, 10 evaluators reviewed the notes, reporting that 75% of the notes were clinically usable without manual corrections. Additionally, 89% of the evaluators believed that implementing MediGen in hospitals could significantly reduce the administrative workload on physicians, improving both healthcare efficiency and physician well-being.

### *B. Ablation Studies*

To gain a deeper understanding of the contributions of model architecture and fine-tuning techniques, an ablation study was conducted to assess the impact of model selection and instruction tuning. Specifically, we fine-tuned similar models like Mistral-7B [9], Gemma-7B [8], and Phi-3-mini-4k-instru-ft using consistent evaluation settings to compare their performance against LLaMA3-8b-ft, aiming to determine the reason behind LLaMA3's superior performance.

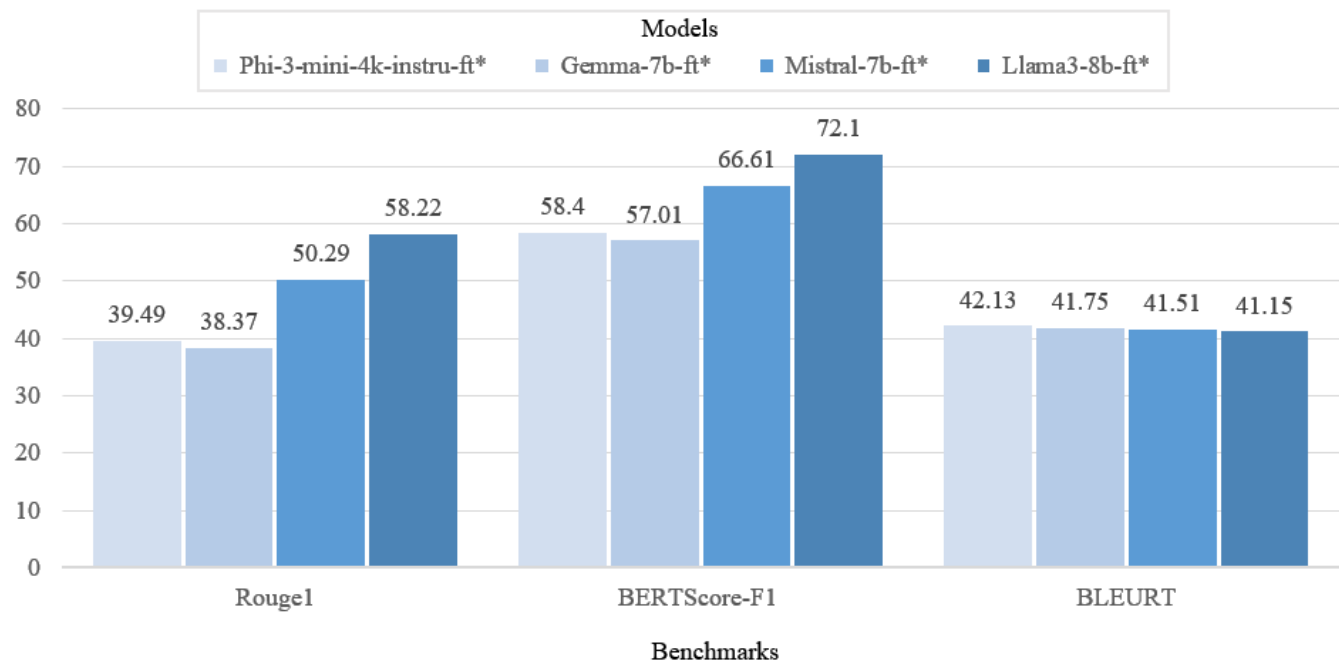

Fig 3. Impact of Model Selection on Performance.

As depicted in Figure 3, each bar illustrates the performance of different models on several benchmarks, with instruction tuning and fine-tuning techniques applied. **LLaMA3-8B-ft** was the top performer, excelling in capturing both semantic richness and clinical relevance, with a BERTScore-F1 of 72.1 and ROUGE-1 of 58.22. It effectively generated accurate medical summaries by ensuring both content precision and coherence. While ROUGE-1 highlighted content overlap, LLaMA3-8B-ft stood out for its ability to grasp deeper clinical meaning. **Mistral-7B-ft** performed well with structured data, particularly in resource-limited environments where its efficiency shone. However, it struggled with semantic coherence in nuanced parts of medical documentation. Its strength lies in handling objective clinical data but lacks depth in more complex contexts.

**GEMMA-7B-ft** showed solid performance in conversational summarization, especially in question-and-answer interactions. Nevertheless, its BERTScore and ROUGE values were lower compared to LLaMA3-8B-ft, particularly in technical areas. It works well for conversational tasks but falls short with more structured medical content. **Phi-3-mini-4k-instru-ft** excelled in niche medical applications, displaying strong performance in specific domains. It slightly outperformed the other models in BLEURT, indicating better capability in generating nuanced text. However, it lagged in BERTScore and ROUGE, making it ideal for domain-specific tasks but less effective in generalized use cases.

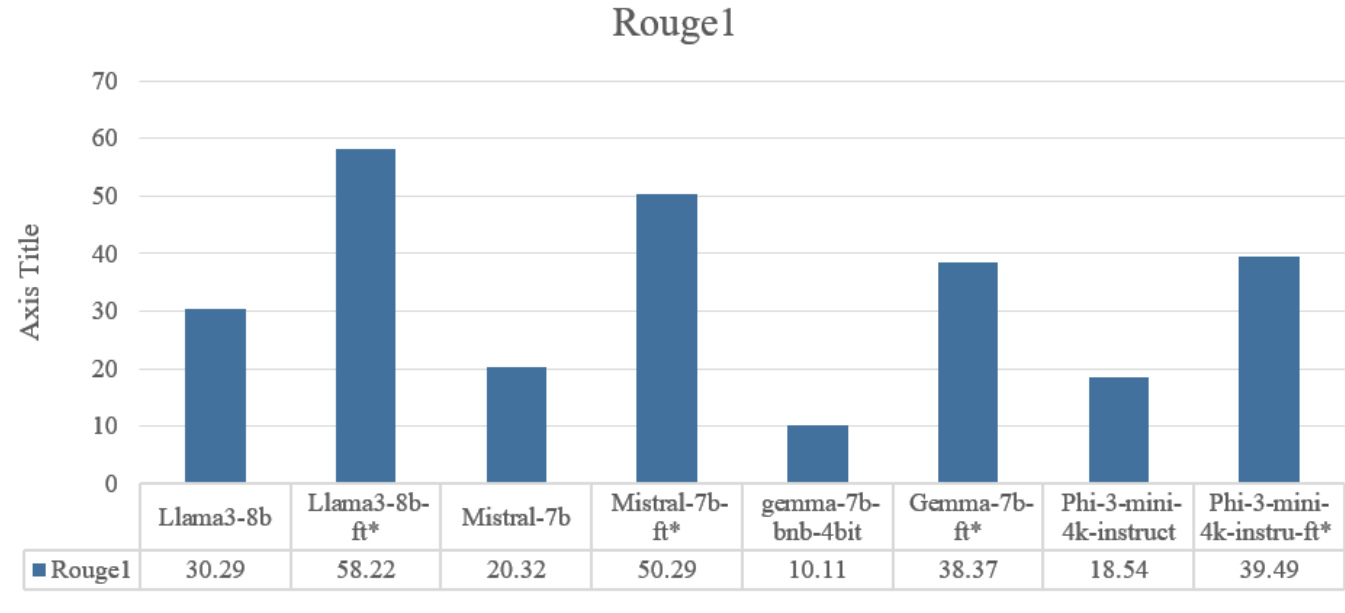

Fig 4. Impact of Instruction Tuning on ROUGE-1 Performance.

Figure 4 highlights the role of instruction tuning in improving performance on the ROUGE-1 benchmark. The ablation study confirms that instruction tuning is crucial for boosting both accuracy and completeness in generated summaries. For instance, when instruction tuning is removed, performance degrades significantly, leading to less structured and incomplete reports. Specifically, models like Gemma-7B-ft and Phi-3-mini-4k-instru-ft without instruction tuning perform notably worse compared to their fine-tuned counterparts, indicating that instruction tuning contributes directly to the quality and structure of the output.

## V. DISCUSSION

### A. *Limitations*

The dataset used to train the MediGen model has inherent limitations due to the sensitive nature of medical information and strict privacy regulations, such as HIPAA in the U.S. Obtaining real clinical encounter data, particularly recordings of patient-doctor conversations, is challenging because sharing such data could compromise patient confidentiality. As a result, the dataset is relatively small in scale and may consist of synthetic or anonymized data. While this approach allows for the development of a model that performs well in clinical dialogue and note generation, it may not fully capture the diversity and complexity of real-world medical encounters.

The limited size and scope of the dataset mean it may not cover the **full spectrum of clinical settings**. For instance, since the dataset primarily includes encounters from general outpatient clinics, the model's ability to generalize to other environments, such as emergency departments or specialist consultations, could be restricted. This limitation can lead to reduced performance when the model is deployed in more diverse healthcare settings. To improve the model's generalization across different contexts, it will be crucial to include a broader range of clinical scenarios in future datasets.

Another challenge lies in **condition-specific generalization**. Clinical dialogues and note formats can vary significantly depending on the medical condition. While the dataset may cover common primary care scenarios, such as joint pain, back pain, and minor injuries, more complex conditions like chronic diseases or rare disorders may be underrepresented. Without sufficient examples of these more intricate cases, the model may struggle to generalize effectively when handling such scenarios.

### B. *Application*

In the Intensive Care Units (ICU), where precise and swift documentation is crucial, MediGen can automatically generate real-time medical reports by transcribing clinician-patient interactions and monitoring vital signs. Integrated with bedside monitors or voice-activated devices, MediGen can continuously update patient records with minimal manual input, allowing healthcare providers to focus more on patient care. This streamlines the documentation process, ensuring timely and accurate reporting of patient conditions and treatments. However, implementing MediGen in the ICU poses challenges, such as ensuring real-time data processing without delays and addressing ethical concerns around continuous voice recording in sensitive clinical environments. Hospitals would need to establish clear consent protocols and guidelines for its use, balancing the need for efficiency with privacy considerations for both patients and healthcare professionals.

In outpatient clinics and telemedicine, MediGen can help manage high patient volumes by automatically generating concise and accurate summaries after consultations. For telemedicine, it could integrate with existing video conferencing platforms to transcribe and summarize conversations between physicians and patients, reducing the documentation workload and improving the efficiency of virtual visits. These summaries could then be directly entered into the EHR for easy access and follow-up care. Challenges in these settings include ensuring seamless integration with various telemedicine platforms and maintaining the accuracy of speech recognition, particularly when dealing with diverse accents and medical terminology. Despite these hurdles, MediGen has the potential to significantly improve the efficiency and accuracy of documentation in both critical care and outpatient settings, benefiting clinicians and patients alike.

### C. Ethical Considerations

The integration of AI into medical documentation brings immense potential, but it also raises crucial ethical considerations, particularly around data privacy and bias. In handling sensitive medical data, AI systems like MediGen must comply with strict regulations such as HIPAA, ensuring robust encryption, anonymization, and clear consent protocols. Patients need to be fully informed about how their data will be used, particularly when recordings of doctor-patient conversations are involved. Additionally, long-term data storage and access raise important questions about retention policies and the ongoing protection of patient confidentiality.

Model bias presents another significant challenge, as AI systems trained on incomplete or unrepresentative datasets may struggle to generalize across diverse populations and clinical settings. If the training data lacks representation of certain groups—whether by ethnicity, geography, or specific medical conditions—the AI may generate less accurate notes, potentially leading to disparities in care. To mitigate bias, it is essential that the dataset is diverse and that the model's performance is continuously monitored across all patient demographics. Regular audits and adjustments are necessary to ensure that AI-generated documentation remains equitable and reliable for all patients.

## VI. CONCLUSIONS

In this study, we introduced MediGen, a fine-tuned large language model designed to automate the generation of medical reports from doctor-patient dialogues, thereby addressing the administrative burden on healthcare professionals. By utilizing state-of-the-art models like LLaMA3-8B, GEMMA-7B, and Mistral-7B, combined with advanced fine-tuning techniques such as Quantized Low-Rank Adaptation (QLoRA) and Parameter-Efficient Fine-Tuning (PEFT), we demonstrated significant improvements in the accuracy, structural coherence, and semantic relevance of generated medical reports. Our results, validated through metrics like ROUGE and BERTScore, confirmed the superior performance of fine-tuned models compared to baseline approaches. Furthermore, the ablation study underscored the importance of instruction tuning in enhancing both the accuracy and completeness of the generated reports.

While MediGen shows strong potential in automating medical documentation, there are clear paths for further improvement. One key step is expanding the diversity of the training dataset. Incorporating data from a wider range of medical scenarios, including emergency rooms and specialist consultations, as well as more varied medical conditions like chronic diseases and rare disorders, would enhance the model's ability to generalize across different clinical settings. Data augmentation techniques, such as dialogue transcription and synthetic data generation, can further increase dataset diversity, making the model more adaptable to real-world interactions. These methods would help MediGen handle a variety of conversation styles, accents, and medical jargon with greater accuracy. Integrating continuous feedback from healthcare professionals is also crucial. Regular assessments from clinicians can identify areas for improvement in accuracy and clinical relevance. Fine-tuning the model through approaches like reinforcement learning from human feedback (RLHF) could further align MediGen with the practical needs of medical professionals, ensuring its ongoing evolution and effectiveness in real-world healthcare environments.